\documentclass{article}

\usepackage{latexsym}
\usepackage{graphics}   
\usepackage{epsf}
\newlength{\largeurtexte}
\newcommand{\omitted}[1]{}
\newcommand{\cause}{\ensuremath{~causes~}}

\newcommand{\isa}{\ensuremath{\rightarrow_{IS-A}}}
\newcommand{\explsf}[2]{\ensuremath \;\textit{explains}\;#1\;\textit{bec\_poss}\;#2 }

\begin{document}
\title{A formalism for causal explanations with an Answer Set Programming translation
\vspace{2ex}\\\protect\small {\rm HAL version, Aug 23, 2010 see LNCS (KSEM 2010) for conference version.}}

\author{Yves Moinard\\
{\small INRIA Bretagne Atlantique, IRISA, 
Campus de Beaulieu, 35042 Rennes cedex, France}\\
{\small email: moinard@irisa.fr }}
\date{\today}
\bibliographystyle{plain}
\maketitle

\begin{abstract}
We examine the practicality for a user of using Answer Set Programming (ASP) for representing logical formalisms.
Our example is a formalism aiming at capturing causal explanations from causal 
information. We show the naturalness and relative efficiency of this translation job.
We are interested in the ease for writing an ASP program. Limitations of the earlier systems  made that in practice, the ``declarative aspect'' was more theoretical than practical. We show how recent improvements in working ASP systems
facilitate the translation.\protect\vspace{-1ex}
\end{abstract}
\section{Introduction}
We consider a formalism designed in collaboration with Philippe Besnard
and Marie-Odile Cordier, aiming at a logical formalization of explanations from causal and ``is-a'' statements. Given some information such as ``fire causes smoke'' and ``grey smoke is a smoke'', if ``grey smoke'' is established, we want to infer that ``fire'' is a (tentative) explanation for this fact.
The formalization \cite{BCM08} is expressed in terms of {\em rules} 
such as ``if $\alpha \cause \beta$ and $\gamma \  is a \ \beta$, then
$\alpha \ explains \ \gamma \\ provided \ \{\alpha,\gamma\} \ is\ possible$''.
This concerns looking for paths in a graph and ASP is good for this.
There exists efficient systems,
such as DLV  \cite{LPFEGPS06} or clasp[D] ({\tt www.dbai.tuwien.ac.at/proj/dlv/} or  
{\tt potassco.sourceforge.net/}).

Transforming formal rules into an ASP program is easy.
ASP should then be an interesting tool for researchers when designing a new theoretical
formalization as the one examined here.
When defining a theory, ASP programs should help examining a great number of middle sized examples. 
Then if middle sized programs could work, 
a few more optimization techniques could make real sized examples work with the final theory.

In fact, even if ASP allows such direct and efficient translation, 
a few problems complicate the task.
The poor data types available in pure ASP systems is a real drawback, since
our rules involve sets.  
Part of this difficulty comes from a second drawback:
In ASP, it is hard to reuse portions of a program. 
Similar rules should be written again, in a different way.
Also, ``brave'' or ``cautious'' reasoning is generally not allowed (except with respect to precise ``queries'').
In ASP, ``the problem is the program'' and the ``solution''
consists in one or several sets of atoms satisfying the problem.
Each such set is an {\em answer set}. 
Brave and cautious solutions mean to look for atoms true respectively in some or in 
all the answer sets. 

This complicates the use of ASP:
any modification becomes complex.
However, things are evolving, e.g. DLV-Complex ({\tt www.mat.unical.it/dlv-complex}), deals with the data structure problem and 
DLT \cite{CI06} 
allows the use of ``templates'', convenient for reusing part of a program. 
We present the explanation formalism,
then its ASP translation in DLV-Complex, and we conclude by a few reasonable expectations about
the future ASP systems which could help a final user.

\section{The causal explanation formalism}

\subsection{Preliminaries
{\rm(propositional version, cf \cite{BCM08} for 
the full formalism).}}\label{subexplic}
We distinguish various types of statements:
\begin{itemize}
\item[$C$:]
A theory expressing causal statements. E.g. $On\_alarm \cause
H\!eard\_bell$.
\item[$O$:]
$IS$-$A$ links between items
which can appear in a causal statement.E.g.,\\ 
$Temperature\_39 \isa Fever\_Temperature$,\\ 
$Heard\_soft\_bell \isa Heard\_bell$.
\item[$W$:]
A classical propositional theory expressing truths 
(incompatible facts, co-occurring facts, $\ldots$). E.g.,
$Heard\_soft\_bell \rightarrow \neg Heard\_loud\_bell$.
\end{itemize}
Propositional symbols denote states of affairs, which can be ``facts'' or ``events''  
such as $Fever\_Temperature$ or $On\_alarm$.
The causal statements express causal relations between facts or events.

Some care is necessary when providing these causal and ontological atoms.
If ``$Flu \cause Fever\_Temperature$'', we conclude
{\em $Flu$ explains $Temperature\_39$} from 
$Temperature\_39 \isa Fever\_Temperature$, but we cannot 
state\\
 $Flu \cause Temperature\_39$: the causal information must be
``on the right level''.

The formal system infers formulas denoting explanations
from $C\cup O \cup W$.
The IS\_A  atoms express knowledge necessary
to infer explanations.
In the following, $\alpha,\beta, \ldots$ denote the propositional 
atoms and $\Phi, \Psi, \ldots$ denote sets thereof.\vspace{1ex}

\textbf{Atoms}\vspace{-1ex}
\begin{enumerate}
\item \emph{Propositional atoms}: $\alpha,\beta, \ldots$.
\item \emph{Causal atoms}: $\alpha \cause \beta$.
\item \emph{Ontological atoms}: $\alpha \isa \beta$.\hspace{2em} 
Reads: $\alpha\mbox{\emph{ is a }} \beta$.
\item \emph{Explanation atoms}: 
$\alpha \explsf{\beta}{\Phi}$. 
Reads: \emph{$\alpha$ is an explanation for $\beta$ because $\Phi$ is possible.}
\end{enumerate}

\textbf{Formulas}\vspace{-1ex}
\begin{enumerate}
  \item \emph{Propositional formulas}: Boolean combinations of propositional
        atoms. 
  \item \emph{Causal formulas}: 
         Boolean combinations of causal or propositional atoms.
\end{enumerate}

The premises $C \cup O \cup W$
consist of propositional and causal formulas, and
ontological \emph{atoms} (no ontological formula), without explanation atom.

\begin{enumerate}
\item\label{propcausprop0} \textbf{Properties of the causal operator} 
  \begin{enumerate}
  \item \label{proofimplprop0} \emph{Entailing [standard] implication}:
      If $\alpha \cause \beta$, then $\alpha \rightarrow \beta.$
  \end{enumerate}
\item \label{propontprop0} \textbf{Properties of the ontological operator}
  \begin{enumerate}
  \item \label{ontoimplprop0} \emph{Entailing implication}:
      If $\alpha \isa \beta$, then $\alpha \rightarrow \beta.$
  \item\label{ontotransprop0} \emph{Transitivity}:
     If $a \isa b$ and $b \isa c$, then $a \isa c$.
  \item\label{ontorefprop0} \emph{Reflexivity}:
     $c \isa c$.\hfill (unconventional, keeps the number of rules low).
  \end{enumerate}
\end{enumerate}

\subsection{The formal system }\label{proofsystem}

\begin{enumerate}
\item\label{propcauschema0} 
\textbf{\emph{Causal atoms entail implication:}} 
     $(\alpha \cause \beta) \rightarrow (\alpha \rightarrow \beta)$.
\item\label{propontschema0} \textbf{\emph{Ontological atoms}}
  \begin{enumerate}
  \item\label{ontoimplicschema0} entail implication:
     If $\beta \isa \gamma$ then $\beta \rightarrow \gamma$.
  \item\label{ontotranschema0}transitivity: 
     If $\alpha \isa \beta$ and $\beta \isa \gamma$ then $\alpha \isa \gamma$.
  \item\label{ontorefschema0} reflexivity:
     $\alpha \isa \alpha$\\\mbox{~}
  \end{enumerate}
\item\label{explicprop5} \textbf{\emph{Generating the explanation atoms}}%
  \begin{enumerate}
 \item\label{explicontdnupprop5} {\em Initial case} \ 
   If  $\delta \isa  \beta, \hspace{1em} \delta \isa \gamma$, and
   $ W \not\models \neg (\alpha \wedge \delta)$,\\\makebox[2em]{}
   then \hspace{1.5em}  $(\alpha \cause \beta) \hspace{1.5em}  \rightarrow \hspace{1.5em} 
        \alpha \explsf{\gamma}{\{\alpha,\delta\}.}$%
  \item\label{transexplicprop5} {\em Transitivity (gathering the
      conditions)}\hspace{1em}
      If\hspace{1em} $W \not \models \neg \bigwedge (\Phi \cup\Psi)$,\\
       then \makebox[1em]{} $(\alpha \explsf{\beta}{\Phi} \;\;\wedge \;\;  
                  \beta \explsf{\gamma}{\Psi}) \makebox[1em]{}  \rightarrow\\  
       \makebox[3.5em]{} \alpha \explsf{\gamma}{(\Phi \cup \Psi)}.$
  \item\label{simplexplicprop5} {\em Simplification of the set of conditions}
      \hfill If 
      $W \models \bigwedge\Phi \rightarrow \bigvee_{i=1}^n \bigwedge\Phi_i$,
      \hfill then\\\makebox[1em]{} 
  $\bigwedge_{i \in \{1,\cdots,n\}} \alpha \explsf{\beta}{(\Phi_i \cup \Phi)}$
       \hspace{.5em}   $\rightarrow$ \hspace{.5em} 
          $\alpha \explsf{\beta}{\Phi}.$
  \end{enumerate}
\end{enumerate}

The elementary ``initial case''
applies (\ref{ontorefschema0}) upon
(\ref{explicontdnupprop5}) where $\beta=\gamma=\delta$,
together with a simplification (\ref{simplexplicprop5}) 
since $\alpha \rightarrow \beta$ here, getting:

If $\alpha \cause  \beta$ and  $W \not\models \neg \alpha$ then 
 $\alpha \explsf{\beta}{\{\alpha\}}$.

These rules are intended as a compromise between expressive power, naturalness of description and relatively efficiency.

Transitivity of {\em explanations} occurs
(gathering conditions).
The simplification rule (\ref{simplexplicprop5})
is powerful
and costly, so the
ASP translation implements the following weaker rule
(also, it never removes $\{\alpha\}$ from $\Phi$):
[\ref{simplexplicprop5}'] 
If $W \models \bigwedge\Phi - \{\varphi\}  \rightarrow \bigwedge\Phi$,
        and $\alpha \explsf{\beta}{\Phi}$ then
        $\alpha \explsf{\beta}{\Phi- \{\varphi\} }.$

An atom $\alpha \explsf{\beta}{\Phi}$ is {\em optimal} if there is no 
explanation atom
$\alpha \explsf{\beta}{\Psi}$ where $W\models \bigwedge\Psi \rightarrow \bigwedge\Phi$ an not conversely.
Keeping only these weakest sets of conditions
is useful when the derivation is made only thanks to the part of $W$ coming from Points \ref{propcauschema0} and \ref{ontoimplicschema0} above.
This keeps all the relevant explanation atoms 
and is easier to read.

\subsection{A generic diagram}\label{subdiagram}

The following diagram summarizes many patterns of inferred explanations:

\noindent$\alpha \cause \beta$, \hfill 
$\alpha \cause \beta_0$, \hfill 
$\beta_2 \cause \gamma$, \hfill 
$\beta_1 \cause \gamma$, \hfill $\,$\\
$\beta_3 \cause \epsilon$, \hfill 
$\gamma_1 \cause \delta$,  \hfill 
$\gamma_3 \cause \delta$, \hfill $\epsilon_3 \cause \gamma_3$; \hfill $\,$\\ 
$\beta \isa \beta_2$, \hfill $\beta_1 \isa \beta$, \hfill  
$\beta_3 \isa \beta_0$, \hfill $\beta_3 \isa \beta_1$, \hfill $\,$\\
$\gamma_1 \isa \gamma$, \hfill $\gamma_2 \isa \gamma$, \hfill 
$\gamma_2 \isa \gamma_3$, \hfill $\gamma_2 \isa \epsilon$, \hfill $\,$\\
$\epsilon_1 \isa \epsilon$, \hfill $\epsilon_2 \isa \epsilon$, \hfill 
$\epsilon_1 \isa \epsilon_3$, \hfill $\epsilon_2\isa \epsilon_3$. \hfill $\,$

This example shows various different ``explaining paths'' from a few given
causal and ontological atoms. 
As a first ``explaining path'' from $\alpha$ to $\delta$
we get successively (path (1a):
$\alpha \explsf{\beta_2}{\{\alpha\}}$,\  
$\alpha \explsf{\gamma_1}{\{\alpha,\gamma_1\}}$,
 and $\alpha \explsf{\delta}{\{\alpha,\gamma_1\}}$, giving
$\alpha \explsf{\delta}{\{\alpha, \gamma_1\}}$ (1a).
The four optimal paths from $\alpha$ to $\delta$ are depicted, and non optimal paths 
(e.g. going through $\beta_1$) exist also.

\begin{figure}[h]
\includegraphics{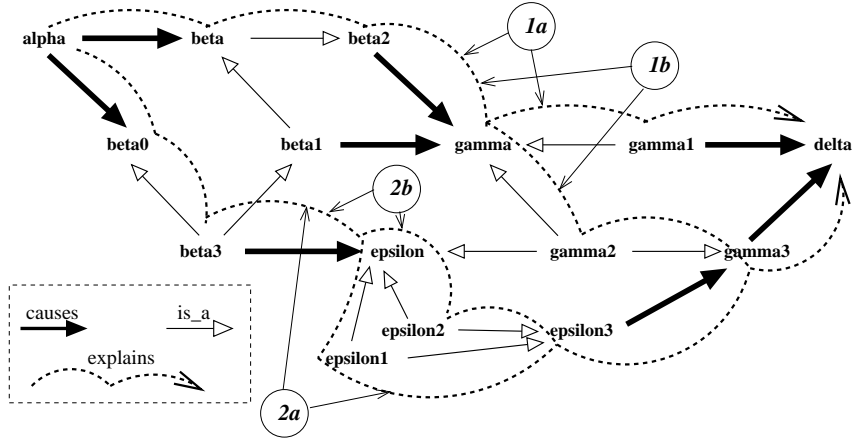}\caption{Four optimal explanation paths from ``alpha'' to ``delta''}  %
\end{figure}%

\section{An ASP translation of the formalism}

\subsection{Presentation}
We describe a program in DLV-Complex. A first version \cite{Moi07} used pure DLV, and was much slower and harder to read.
We have successfully tested an example with more than a hundred symbols and more than 10 
different explanation atoms for some $(I,J)$ (made from two copies of the example
of the diagram, linked through a few more data).
We have encountered a problem, not listed in the ``three problems'' evoked above.
The full program, including ``optimization of the result'', did not work on our computer.
The simplification step and the verification step are clearly separated from the first generating step, thus we have split the program in three parts:
The first one generates various explanation atoms (including all the optimal ones).
The second program keeps only the optimal explanation atoms, 
in order to help reading the set of the solutions, following Point \ref{simplexplicprop5}'.
The third program  checks whether the set of conditions is satisfied 
in the answer set considered. Then our ``large example'' works.
Notice that it useful to enumerate the answer sets (as described in the simple and very interesting \cite{TBA05}) in order to help this splitting. 

\subsection{The generating part: getting the relevant explanation atoms}\label{subgen}

The ``answer'' of an ASP program is a set of {\em answer sets}, that is a set of 
concrete literals 
satisfying the rules (see e. g. \cite{Bar03} for exact definitions
and \cite{LPFEGPS06} for the precise syntax of DLV ({\sf ``:-''} represents ``$\leftarrow$'' and 
{\sf ``-''} alone represents ``standard negation'' while {\sf ``v''} is the disjonction symbol in 
the head of the rules, and {\sf ``!=''} is ``$\not =$'').\\
The user provides the following data:

{\sf symbol(alpha)} for each propositional symbol {\sf alpha}.\\\indent
{\sf cause(alpha,beta)} for each causal atom $\alpha \cause \beta$,\\\indent
{\sf ont(alpha,beta)} for each ``is\_a'' atom $\alpha \isa \beta$ and \\\indent
{\sf true(alpha)} for each other propositional atom $\alpha$ involved in formulas.

Causal and propositional formulas, such as 
$ (\neg\epsilon1 \wedge \neg \epsilon2) \vee \neg \gamma1 \vee \neg \gamma2$  must be put in conjunctive normal form, in order to be entered as sets of clauses:\\
\indent \{\ {\sf -true(epsilon1) v -true(gamma1) v -true(gamma2).}\ ,\\
\indent \makebox[.5em]{} {\sf -true(epsilon2) v -true(gamma1) v -true(gamma2).\ }\}  

The interesting result consists in the explanation predicates:

{\sf ecSet(alpha,beta,\{alpha,delta,gamma\})} represents the explanation atom

$\alpha \explsf{\beta}{\{\alpha,\delta\}}$.

Here come the first rules (cf \ref{ontotranschema0} \S \ref{proofsystem}):\\
{\sf ontt(I,J) :- ont(I,J). \ ontt(I,K) :- ontt(I,J), ont(J,K).
     \ ontt(I,I) :- symbole(I).}

We refer the reader to \cite{Moi10A} for more details about the program.
As an example of the advantage of using sets, let us give here the rule dealing with transitivity of explanations (cf \ref{transexplicprop5} \S \ref{proofsystem}),
{\sf ecinit} referring to explanations not using transitivity rule.
({\sf \#insert(Set1,E2,Set)} means: $Set = Set1 \cup \{E2\}$):

{\sf ecSet(I,J,\{I\}) :- ecinit(I,J,I).
ecSet(I,J,\{I,E\}) :- ecinit(I,J,E), not ecSet(I,J,\{I\}).}

{\sf ecSet(I,J,Set) :- ecSet(I,K,Set1), not ecSet(I,J,Set1), ecinit(K,J,E2), E2 != K,\\\makebox[8em]{} \#insert(Set1,E2,Set).

ecSet(I,J,Set) :- ecSet(I,K,Set), ecinit(K,J,K).}

\subsection{Optimizing the explanation atoms}\label{subopt}

The ``weak simplification'' rule \ref{simplexplicprop5}' \S \ref{proofsystem}) 
is used at this step, omitted here for lack of space. Moreover,
if two sets of condition $\Phi$, $\Psi$ exist for some $\alpha \ explains \ \beta$, and if
$\Phi\models \Psi$ and not conversely, then only $\alpha\explsf{\beta}{\Psi}$ is kept, the stronger set $\Phi$ being discarded.
This avoids clearly unnecessary explanation atoms. 
This part (not logically necessary) is costly, but helps interpreting the result by a human reader.

\subsection{Checking the set of conditions}\label{subckeck}

Finally, the following program starts from the result
of any of the last two preceding programs and checks, in each answer set, whether the set of conditions is satisfied or not.
The result is given by {\sf explVer(I,J,Set)}: $I \explsf{J}{Set}$ where
$Set$ is satisfiable in the answer set considered (``{\sf Ver}'' stands for ``verified'').

{\sf explSuppr(I,J,Set) :- ecSetRes(I,J,Set), -true(E), \#member(E,Set).

explVer(I,J,Set) :- ecSetRes(I,J,Set), not explSuppr(I,J,Set).}

Only ``individual'' checking is made here, in accordance with the requirement that the computational properties remain manageable.

With the whole chain (\S \ref{subgen}, \ref{subopt}, and \ref{subckeck}), 
modifying a rule of the formalism can  be done easily.
The gain of using DLV-Complex instead of pure DLV
(or gringo/claspD) is significant and worth mentioning.

\subsection{Conclusion and future work}
We have shown how the recent versions of running ASP systems allow easy translation of
logical formalisms. The example of the explanation formalism shows that
such a translation can already be useful for testing new theories.
In a near future, cases from the ``real world'' should be manageable
and the end user should be able to use ASP for a great variety of diagnostic problems.

Here are two considerations about what could be hoped for future ASP systems
in order to deal easily with this kind of problem.
Since ASP systems are regularly evolving, we can hope that a near future the annoying trick consisting in launching the programs one after the other, and not in a single launch, should become unnecessary.
It seems easy to detect 
that some predicates can safely be computed first, before launching the subsequent computation. In our example, computing {\sf ecSet} first, then {\sf ecSetRes} and finally {\sf ecSetVer} is possible, and such one way dependencies could be detected. 
The great difference in practice between launching the three programs together, and launching them one after the other, shows that such improvement could have  spectacular consequences. 

Also, efficient ``enumerating meta-predicates'' would be useful (even if logically useless).
A last interesting improvement would concern the possibility of implementing 
``enumerating answer sets'', as described in e.g. \cite{TBA05}, then complex comparisons between answer sets could be made.

For what concerns our own work, the important things to do are to apply the formalism to real situations, and, to this respect, firstly to significantly extend our notion of ``ontology'' towards a real one.

 
\end{document}